\documentclass[10pt,twocolumn,letterpaper]{article}

\usepackage{cvpr}              
\usepackage{xspace}
\usepackage{multirow}
\usepackage{xcolor}
\definecolor{cvprblue}{rgb}{0.21,0.49,0.74}
\usepackage[pagebackref,breaklinks,colorlinks,allcolors=cvprblue]{hyperref}

\def\paperID{****} 
\def\confName{CVPR}
\def\confYear{2026}

\title{ShotDirector: Directorially Controllable Multi-Shot Video Generation with Cinematographic Transitions}

\author{
Xiaoxue Wu$^{1,2*}$~~~~
Xinyuan Chen$^{2\dag}$~~~~ 
Yaohui Wang$^{2}$~~~~
Yu Qiao$^{2\dag}$~~~~\\
\\
$^{1}$Fudan University~~~~~~~~~~
$^{2}$Shanghai Artificial Intelligence Laboratory \\
}

\def\methodname{ShotDirector\xspace}
\def\datasetname{ShotWeaver40K\xspace}

\begin{document}
\twocolumn[{
\renewcommand\twocolumn[1][t!]{#1}%
\maketitle

\begin{center}
    \centering
    \includegraphics[width=1.0\linewidth]{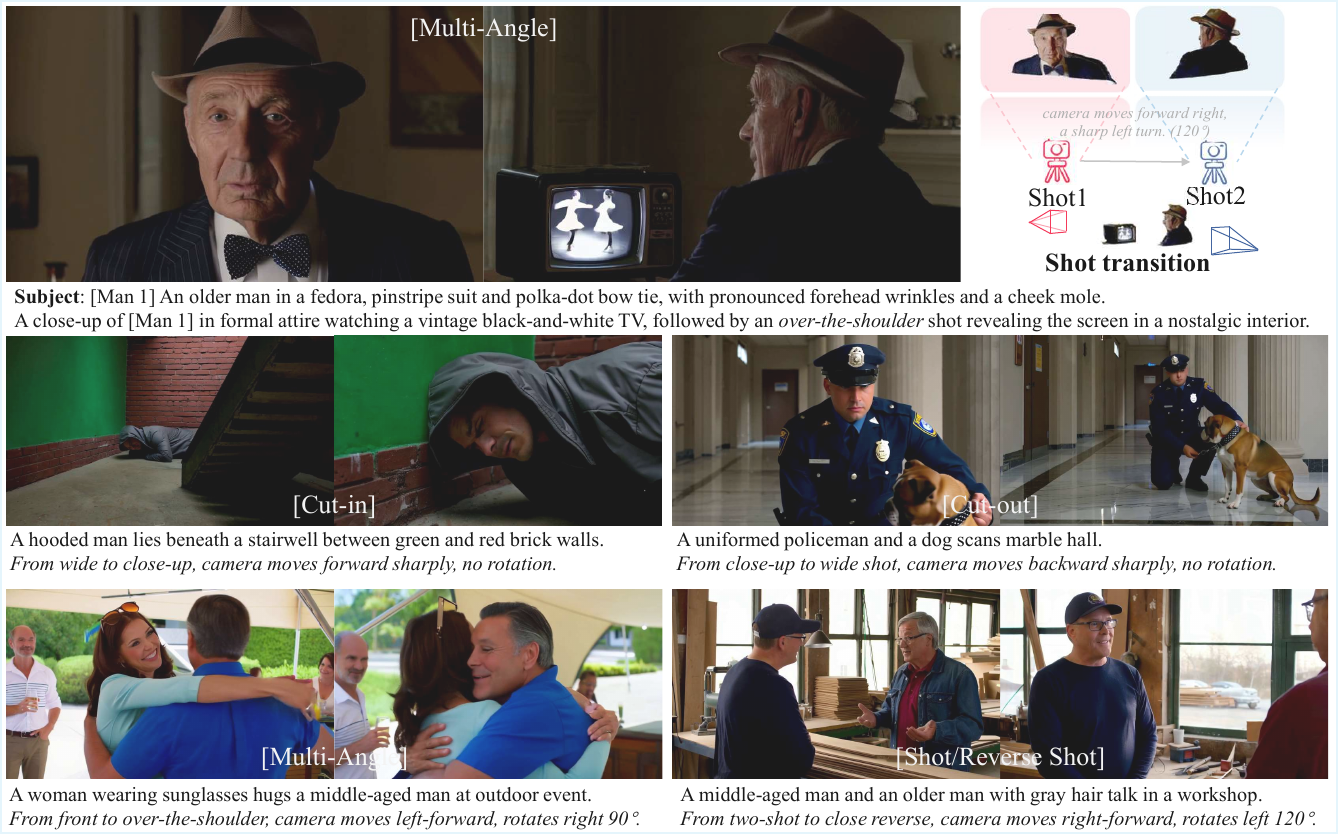}
   \captionof{figure}{Multi-shot videos with professional editing patterns generated by \methodname. We showcase representative cases of different shot transitions, annotated with concise corresponding context and transition control information.}
   \label{fig:teaser}
\end{center}}]

\renewcommand{\thefootnote}{}%
\footnotetext{
\textsuperscript{*}Work done during an internship at Shanghai AI Laboratory.\\  
\hspace*{1.8em}\textsuperscript{\dag}Corresponding authors.}%
\renewcommand{\thefootnote}{\arabic{footnote}}%

\begin{abstract}
Shot transitions play a pivotal role in multi-shot video generation, as they determine the overall narrative expression and the directorial design of visual storytelling.
However, recent progress has primarily focused on low-level visual consistency across shots, neglecting how transitions are designed and how cinematographic language contributes to coherent narrative expression. This often leads to mere sequential shot changes without intentional film-editing patterns. To address this limitation, we propose \textbf{\methodname}, an efficient framework that integrates parameter-level camera control and hierarchical editing-pattern-aware prompting.
Specifically, we adopt a camera control module that incorporates 6-DoF poses and intrinsic settings to enable precise camera information injection. In addition, a shot-aware mask mechanism is employed to introduce hierarchical prompts aware of professional editing patterns, allowing fine-grained control over shot content. Through this design, our framework effectively combines parameter-level conditions with high-level semantic guidance, achieving film-like controllable shot transitions.
To facilitate training and evaluation, we construct \datasetname, a dataset that captures the priors of film-like editing patterns, and develop a set of evaluation metrics for controllable multi-shot video generation. Extensive experiments demonstrate the effectiveness of our framework. Project Page: \url{https://uknowsth.github.io/ShotDirector/}
\end{abstract}    
\section{Introduction}
\label{sec:intro}

\begin{figure*}[!t]
  \centering
   \includegraphics[width=1.0\linewidth]{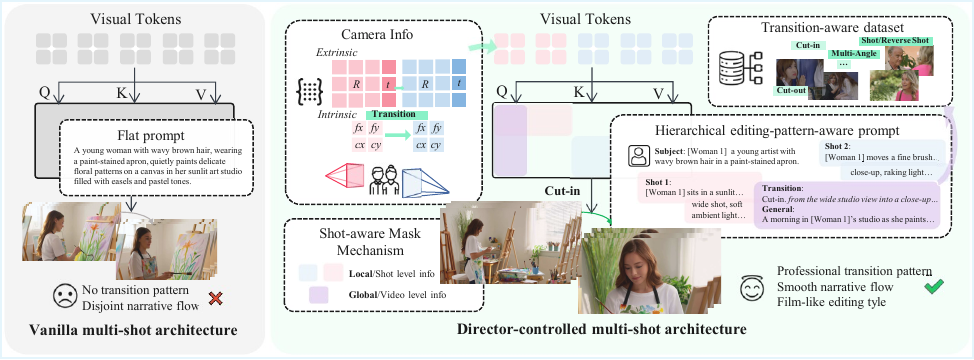}

   \caption{\textbf{Overview of \methodname.}
Conventional end-to-end multi-shot video models use flat prompts, treating shot transitions merely as visual break without explicit design. Our method introduces hierarchical editing-pattern-aware prompting and camera information injection to explicitly model shot transitions, while a shot-aware mask mechanism enables fine-grained control. Trained on the transition-aware dataset \datasetname, our model generates multi-shot videos with professional transition patterns and a smooth narrative flow.}
   \label{fig:overview}
\end{figure*}

Recent advances in diffusion-based \cite{stablevideodiffusion,song2020score} video generation have brought unprecedented progress in synthesizing realistic and temporally coherent videos from textual descriptions \citep{wan2025wan,kong2024hunyuanvideo,li2024longvideosurvey, wang2024lavie}. Building upon architectures such as Diffusion Transformer \cite{DiT} and the latent diffusion framework \cite{ho2020denoising, rombach2022high}, recent models have demonstrated remarkable capabilities in generating photorealistic single-shot videos that exhibit strong spatial and temporal fidelity \cite{sora,wan2025wan,yang2024cogvideox}. Following these successes, research attention has gradually shifted toward multi-shot video generation \cite{chen2024seine,qi2025mask2dit,guo2025longcontexttuning,wu2025cinetrans,ttt}, which aims to synthesize film-like narratives through shot transitions that collectively convey cinematic rhythm and artistic expression.

Existing paradigms for multi-shot video generation fall into two main categories. The stitching-based approaches \cite{zhou2024storydiffusion,zheng2024videogenofthought,long2024videostudio,he2023animateastory,xie2024dreamfactory,zhao2024moviedreamer} synthesize individual shots independently and concatenate them into a sequence. These methods emphasize cross-shot consistency through conditional information injection \cite{long2024videostudio} or keyframe-based alignment \cite{zhou2024storydiffusion}. However, such inter-shot dependencies are externally constrained rather than learned from data, preventing the model from exploiting the priors of film-like editing present in real cinematic corpora. Consequently, the generated sequences resemble collections of single-shot clips, lacking genuine narrative continuity.
More recently, end-to-end diffusion frameworks \cite{guo2025longcontexttuning,qi2025mask2dit,jia2025moga} have been proposed to enable inter-shot interaction within the generative process, yielding higher visual and temporal consistency. Yet, these methods typically regard shot transitions as merely frame-level changes, without explicit modeling of editing conventions (e.g. cut-in, cut-out, shot/reverse-shot) or controllable transition dynamics. As a result, despite their visual fidelity, the produced multi-shot videos remain devoid of cinematic narrative structure and intentional shot design characteristic of professional filmmaking.

In this work, building upon visual coherence, we further explore how shot transitions are intentionally designed to serve cinematic narrative expression. Rather than viewing a cut as an abrupt visual break, we interpret it as a fundamental directorial tool for controlling narrative rhythm and audience perception. For example, shot/reverse shot structures construct dialogue dynamics, while variations in framing and perspective guide emotional focus, as shown in \cref{fig:teaser}. From this standpoint, multi-shot video generation entails not only high-fidelity visual synthesis but also directorial decision-making, i.e., determining how the next shot should unfold. To achieve this, we propose \methodname, a framework that combines parametric camera control with editing-pattern-aware hierarchical prompting, enabling controllable and film-like shot transitions, as illustrated in \cref{fig:overview}.

Specifically, our framework controls the cinematic characteristics of shot transitions from two complementary perspectives: parameter-level camera settings and semantic-level hierarchical prompting, thereby endowing diffusion models with an awareness of professional editing patterns.
First, we regard camera settings as a critical conditioning factor and adopt effective approaches to integrate it into multi-shot video generation. 
We employ Plücker embedding \cite{sitzmann2021light} to encode the geometric information of each pixel’s viewing ray, and integrate it with the direct camera extrinsic parameters through two parallel branches, providing comprehensive representations of camera position and orientation.
This formulation enables refined control over viewpoint shifts and mitigates unintended discontinuities across shots.
Second, we propose a shot-aware mask mechanism that guides context modeling at both global and local levels, enabling fine-grained control over token interactions. By structuring the visibility of tokens, it allows the model to balance global coherence with shot-specific diversity. This mechanism aligns textual information, cinematographic style, and designed shot transitions with corresponding visual tokens, explicitly modeling editing conventions and incorporating directorial priors into the diffusion process.
By integrating parameter-level conditions with high-level semantic guidance, \methodname  produces multi-shot videos that exhibit film-like cinematographic expression and coherent narrative flow.

To train our framework, we construct a high-quality multi-shot video dataset \datasetname. A rigorous data curation pipeline is adopted to carefully filter cinematic raw footage, ensuring that transitions between shots follow plausible narrative or spatial reasoning rather than arbitrary visual changes, thus preventing confusion for the model.  Moreover, cinematography-aware captions provide explicit annotations of editing patterns along with detailed camera parameters for each shot, enabling the model to internalize shot transition design priors and produce professionally edited, film-like multi-shot videos.

For evaluation, we design a comprehensive assessment protocol encompassing controllability, consistency, and visual quality, providing a systematic framework for analyzing the performance of multi-shot video generation. Experimental results demonstrate that \methodname achieves controllable shot transitions that align with specific editing patterns, effectively conforming to intentional film-editing conventions while maintaining high visual fidelity and cross-shot consistency.
Our work establishes a novel perspective on multi-shot video generation, emphasizing the role of directorial control in shot transitions.

\section{Related Work}
\label{sec:related}

\subsection{Multi-shot Video Generation}

With the advancement of diffusion-based video generation methods \cite{xing2024survey,khachatryan2023text2video,zhuang2024vlogger,ma2024latte,wang2024lavie,ho2020denoising,stablevideodiffusion}, models trained on large-scale video-text pair corpora gain initial capability for shot transitions \cite{sora,kong2024hunyuanvideo,wan2025wan}. However, such semantic-driven transitions remain unstable and uncontrollable. To address this, recent studies on multi-shot video generation can be broadly divided into two directions. The first type focuses on shot-by-shot generation \cite{zhao2024moviedreamer,xie2024dreamfactory,he2023animateastory} and emphasises preserving high consistency across shots via external constraints. StoryDiffusion \cite{zhou2024storydiffusion} generates a series of keyframes and then animates them. VideoStudio \cite{long2024videostudio} conditions each storyboard shot on rich metadata. VGoT \cite{zheng2024videogenofthought} uses identity-preserving embeddings to maintain character consistency. While these approaches can indeed generate a continuous video sequence with some consistency, they do not explicitly explore shot transitions themselves. Thus, these works resemble the generation of a sequence of related frames \cite{liu2025phantom,hu2022lora} rather than a multi-shot video with cinematic characteristics.
As the field further developed, end-to-end multi-shot video generation pipelines emerge as a more promising approach \cite{kara2025shotadapter,wu2025cinetrans}. These methods modify conventional diffusion models to allow interaction among different shots within the model.
Mask$^2$DiT \cite{qi2025mask2dit} and CineTrans \cite{wu2025cinetrans} control the occurrence of multi-shot segments through semantic and visual masks, respectively. LCT \cite{guo2025longcontexttuning} employs a specialized positional encoding scheme to facilitate shot-level modeling. MoGA \cite{jia2025moga} introduces customized group attention mechanism that enables the model to handle longer multi-shot sequences effectively, while TTT \cite{ttt} incorporates an additional layer to enhance the model’s capability in managing multi-shot video generation.
Although these methods produce high-quality multi-shot sequences, they still treat shot transitions as abrupt frame changes, lacking controllability and semantic coherence. While Cut2Next \cite{he2025cut2next} considers editing patterns, it remains image-level and thus limited in complex video scenarios.
Our method integrates parameter-level camera control and semantic-level prompting, enabling multi-shot video generation with cinematographic language.

\begin{figure*}
  \centering
  \begin{subfigure}{0.37\linewidth}
    \includegraphics[width=1.0\linewidth]{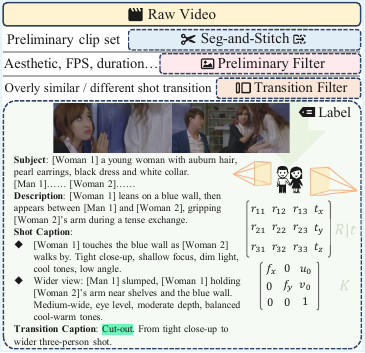}
    \caption{\textbf{Dataset curation pipeline.} Starting from a raw movie dataset, we apply Seg-and-Stitch and filtering to obtain a refined video set, annotated with hierarchical captions and transition camera poses to form \datasetname.}
    \label{fig:dataset}
  \end{subfigure}
  \hfill
  \begin{subfigure}{0.59\linewidth}
    \includegraphics[width=1.0\linewidth]{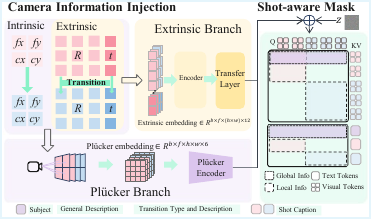}
    \caption{\textbf{The architecture of our method.} We adopt a dual-branch design to inject camera information and employ a shot-aware mask mechanism to regulate token visibility across global and local contexts. Professional transition design is incorporated into global text tokens, working with camera information to achieve multi-granular control over shot transitions.}
    \label{fig:method}
  \end{subfigure}
  \caption{The dataset curation pipeline and model architecture of our framework. The finely annotated dataset is used to train the model, enabling it to learn film-like editing patterns.}
  \label{fig:dataset_and_method}
\end{figure*}

\subsection{Camera-Controlled Video Generation}

Camera-controlled video diffusion models aim to enable explicit control over camera motion during video synthesis. Early approaches \cite{he2024cameractrl,fengi2vcontrol} inject camera parameters into pretrained video diffusion models to achieve viewpoint-aware generation. Subsequent works further incorporate geometric constraints or 3D priors \cite{bahmani2025ac3d,bahmanivd3d,houtraining,xu2024camco}. Beyond single-camera setups, SynCamMaster \cite{baisyncammaster} and ReCamMaster \cite{bai2025recammaster} introduce frameworks for synchronized multi-camera generation and 3D-consistent scene modeling. CameraCtrlII \cite{he2025cameractrl} extends this line of research by enhancing temporal consistency and enabling continuous, controllable camera motion in long video sequences. Our method integrates camera control into the concept of shot transitions, treating camera pose as an essential conditioning factor for multi-shot sequence generation.
\section{Method}
\label{sec:method}

To enable professional shot transitions to serve cinematic narrative expression, we propose the \methodname framework, which determines how the next shot unfolds through parametric camera control and hierarchical editing-pattern-aware prompts, as shown in \cref{fig:dataset_and_method}.  
\cref{sec:dataset} presents \datasetname, a dataset constructed to capture the prior of filmmaking editing patterns.  
\cref{sec:camera} and \cref{sec:mask} describe the camera information injection and the shot-aware mask for fine-grained control of shot transition patterns, respectively.  
Finally, \cref{sec:training} provides additional details.

\subsection{Data Collection}
\label{sec:dataset}

To enable the diffusion model to develop a comprehensive and professional understanding of shot transitions, we construct the dataset through a refined data processing pipeline, as demonstrated in \cref{fig:dataset}. Starting from a collection of raw movie data, we perform shot segmentation \cite{soucek2024transnet} and similar-segment stitching \cite{girdhar2023imagebind} to assemble coherent sequences of shots into video sequences.
During the preliminary filtering stage, indicators such as resolution, frame rate, and aesthetic score ensure high baseline quality, with aesthetic assessment particularly emphasizing adjacent frames around shot transitions for visual clarity.
Moreover, the dataset is further refined by transition quality: shot transition must exhibit sufficient content variation while maintaining causal or spatial continuity, allowing viewers to infer the cut between shots.
Therefore, videos that are overly similar or excessively dissimilar, making the transition logic unintelligible, are filtered out \cite{radford2021learning, wang2024qwen2} to reduce potential confusion. 

In the video captioning stage, we perform hierarchical text annotation and camera pose estimation. 
GPT-5-mini  generates script-level captions for each multi-shot video, covering the subject, overall and shot-wise descriptions, as well as the transition type and its description.
Following \cite{he2025cut2next}, we emphasize four representative transition types commonly used in filmmaking: shot/reverse shot (dialogue-based alternation of perspectives), cut-in (transition to a closer framing of the same subject), cut-out (transition to a wider contextual view), and multi-angle (switching between different viewpoints of the same action), which together provide a clear structure for controllable transition modeling.
Each shot-wise description includes shot content and cinematographic features, enabling the model to capture subject continuity and the semantics of professional camera settings.
For camera pose estimation, VGGT \cite{wang2025vggt} estimates camera rotation and translation relative to the first shot, representing the motion parameters in matrix form. 
Finally, we obtain \datasetname, a high-quality dataset with detailed shot transition annotations, which provides the foundation for training models to generate multi-shot videos aligned with film-editing patterns.

\subsection{Camera Information Injection}
\label{sec:camera}

To enhance controllability of shot transitions in focal center and camera angle, we introduce parametric camera settings into the diffusion model as a critical conditioning signal. 

Specifically, we design a dual-branch architecture that integrates Plücker embedding \cite{sitzmann2021light} and direct camera extrinsic parameters into the denoising process, providing both pixel-wise spatial ray maps and raw camera configurations.  
Conventionally, the camera pose is defined by the intrinsic and extrinsic parameters, denoted as $K \in \mathbb{R}^{3\times3}$ and $E = [R; t] \in \mathbb{R}^{3\times4}$, respectively, where $R \in \mathbb{R}^{3\times3}$ represents the rotation component of the extrinsic parameters and $t \in \mathbb{R}^{3\times1}$ denotes the translation vector.
The extrinsic branch employs an MLP to directly inject the camera extrinsic parameters $E$ into visual latents:
\begin{equation}
C_{\mathrm{extrinsic}}=MLP(\mathrm{flatten}(E))
  \label{eq:C_cam}.
\end{equation}
Despite lacking intrinsic information such as focal length, extrinsic features effectively capture camera orientation cues, which has been validated in \cite{baisyncammaster}.
Furthermore, the Plücker branch follows the conventional formulation to represent the spatial ray map corresponding to the camera information. For each pixel $(u,v)$
 in the image coordinate space, its Plücker representation is defined as follows:
\begin{equation}
    p_{u,v}=(o\times d_{u,v},d_{u,v})\in \mathbb{R}^6,
\end{equation}
where $o \in \mathbb{R}^3$ denotes the camera center in world coordinates, and $d_{u,v} \in \mathbb{R}^3$ is the viewing direction vector from the camera center to pixel $(u,v)$, which is computed as
\begin{equation}
    d_{u,v} = RK^{-1}[u,v,1]^T, 
\end{equation}
and then normalized to unit length. On this basis, the Plücker embedding of each frame is processed by convolutional layers and injected into the visual latents:
\begin{equation}
  C_{\text{Plücker}}=Conv(P),
  \label{eq:C_cam_}
\end{equation}
where $P=[p_{u,v}] \in \mathbb{R}^{h \times w \times 6}$ represents the Plücker embedding of a frame. Finally, the dual-branch camera information is added to the visual tokens $z_i$ associated with the $i$-th shot prior to self-attention:
\begin{equation}
    z_i' = z_i + C_{\text{extrinsic},i} + C_{\text{Plücker},i}.
\end{equation}

Incorporating diverse forms of camera information in shot transitions enables the diffusion model to capture the design intent behind camera settings, thereby providing auxiliary cues for controllable multi-shot video generation.

\begin{figure*}[t]
  \centering
   \includegraphics[width=1.0\linewidth]{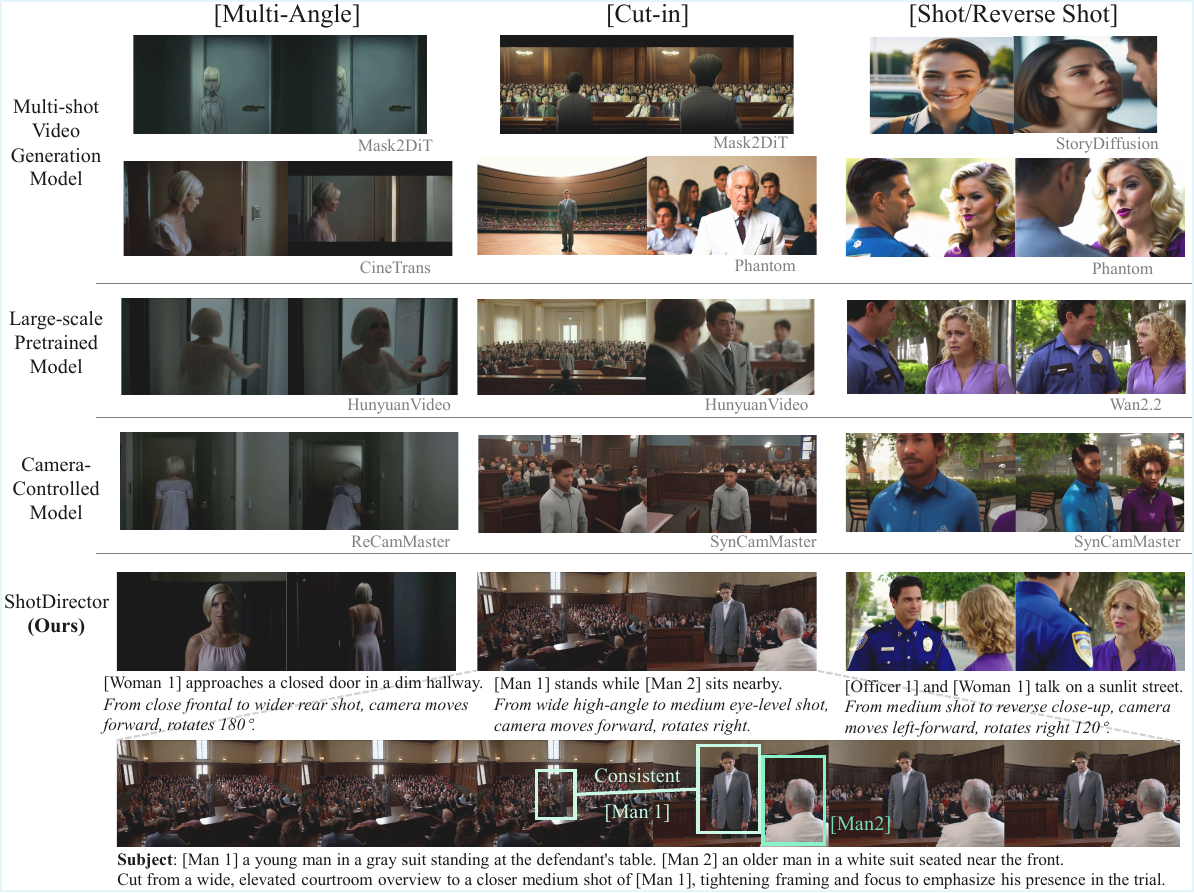}

   \caption{\textbf{Qualitative results for different methods.} Our proposed \methodname generates multi-shot videos with professional shot transition patterns, surpassing various types of existing methods. We presents representative cases from other approaches with shortened prompts for brevity. Please refer to the supplementary project page for more complete results.}
   \label{fig:results}
\end{figure*}

\subsection{Shot-aware Mask Mechanism}
\label{sec:mask}

Beyond parametric conditioning, high-level semantic guidance is essential for the diffusion model to capture professional shot transition patterns. To this end, we adopt a shot-aware mask mechanism that guides each token to incorporate both global and local cues in visual and textual domains, enabling the model to handle hierarchical text captions in a more detailed and structured manner.

Formally, the visual latents $z_i'$ from the $i$-th shot are processed through the attention layers in the DiT architecture, where the shot-aware mask constrains the query to interact only with its corresponding contextual information:
\begin{equation}
\mathrm{Attn}_{\text{shot-aware}}(z_i')=\mathrm{Attn}(q_{z_i'},K^*,V^*),
\end{equation}
where $K^* = [K^{global}_i,K^{local}_i]$ and $V^* = [V^{global}_i,V^{local}_i] $.
Visually, local information refers to all tokens within the current shot, while global information consists of the tokens from the first frame of the entire video, as shown in \cref{fig:method}. To promote sufficient global interaction in early denoising stages, all tokens are kept visible in the initial layers. This mechanism enables each shot to capture overall scene context while retaining shot-specific visual details, aligning with the goal of multi-shot video generation: ensuring high-level contextual consistency while preserving visual diversity.
Textually, local information includes shot-specific descriptions and cinematographic cues, whereas global information covers shared subject attributes, overall narrative, and transition semantics. This design enforces precise alignment between textual guidance and corresponding visual tokens. The subject label maintains inter-shot consistency, while transition semantics provide prior knowledge of professional editing patterns, leading to coherent and controllable transitions.
The shot-aware mask facilitates structured interaction between global and local contexts, allowing each shot to stay contextually consistent with the whole video while preserving its distinct appearance, thereby introducing a hierarchical and editing-aware prompting strategy.

\subsection{Implementation} 
\label{sec:training}

To integrate conditioning signals into the denoising process, we perform training based on \cite{wan2025wan}. The extrinsic branch is initialized with the camera encoder from \cite{bai2025recammaster}, followed by a zero-initialized MLP transfer layer connecting to the DiT framework, while the Plücker branch is randomly initialized. During the warm-up phase, only the dual-branch encoder is trained, with the extrinsic branch limited to its transfer layer. The self-attention parameters are subsequently unfrozen for joint optimization.  
Since camera poses in real data are less reliable than in synthetic datasets, we adopt a two-stage training scheme. In the first stage, the model is trained on \datasetname. In the second stage, training data are augmented with SynCamVideo \cite{baisyncammaster} at a 7:3 real-to-synthetic ratio. This two-stage strategy enables the model to learn transition behavior and leverage camera information as auxiliary guidance for stable and controllable multi-shot video generation.

\section{Experiment}
\label{sec:experiment}

This section presents the experimental setup, evaluation protocols, and results. \cref{sec:implementation details} outlines the parameter settings, training configurations and baselines used in our experiments. \cref{sec:evaluation} and \cref{sec:ablation} describe the evaluation setup and results, demonstrating that the proposed components effectively enhance multi-shot video generation and outperform existing baselines. Finally, \cref{sec:addition} introduces further extensions and capabilities of our model.

\subsection{Experimental Setup}
\label{sec:implementation details}

\noindent \textbf{Implementation Details}.  
We adopt Wan2.1-T2V-1.3B \cite{wan2025wan} as the base model and train it 
on NVIDIA A800 GPUs.  
Following the training scheme described in \cref{sec:training}, in the first stage, we use the Adam optimizer \cite{adam2014method} with a learning rate of $1 \times 10^{-4}$ for 10,000 steps, enabling the model to initially learn transition capabilities.  
In the second stage, the learning rate is set to $5 \times 10^{-5}$ , and the model is trained for 3,000 steps, enhancing its understanding of transition design.

\noindent \textbf{Baselines}.
We compare our method against three categories of strong baselines. 
For multi-shot video generation, we adopt Mask$^2$DiT \cite{qi2025mask2dit} and CineTrans \cite{wu2025cinetrans} as end-to-end generation pipelines, and StoryDiffusion \cite{zhou2024storydiffusion} with CogVideoXI2V \cite{yang2024cogvideox} as a shot-by-shot generation approach.
We further include the reference-to-video method Phantom \cite{liu2025phantom}, which utilizes reference images generated by a text-to-image model to compose multi-shot videos.
For pre-trained video diffusion models, we compare against HunyuanVideo \cite{kong2024hunyuanvideo} and Wan2.2 \cite{wan2025wan}.  
Finally, for multi-view and camera-control methods, we evaluate SynCamMaster \cite{baisyncammaster}, which synthesizes two-view videos to produce shot transitions, and ReCamMaster \cite{bai2025recammaster}, which performs camera-controlled video editing based on videos generated by Wan2.2.

\begin{table*}[tb]
  \scriptsize
  \caption{\textbf{Quantitative results}. The best and runner-up are in \textbf{bold} and \underline{underlined}.}
  \label{tab:evaluation}
  \centering
  \renewcommand{\arraystretch}{1.2}%
  \begin{tabular}{l|cc|cccc|cc} 
    \toprule
    \multirow{2}{*}{\parbox{3.5cm}{Method}} & \multicolumn{2}{c}{ \textbf{Transition Control}} & \multicolumn{4}{c}{\textbf{Overall Quality}} & \multicolumn{2}{c}{\textbf{Cross-shot Consistency}} \\ \cline{2-9}

    & Confidence\textuparrow & Type Acc\textuparrow & Aesthetic\textuparrow & Imaging\textuparrow & Overall Consistency\textuparrow & FVD\textdownarrow & Semantic\textuparrow & Visual\textuparrow \\ \hline
    
   Mask$^2$DiT \cite{qi2025mask2dit} & 0.2233 & 0.2033 & 0.5958 & 0.6841 & 0.2184 & 69.49 & 0.7801 & 0.7779 \\
   CineTrans \cite{wu2025cinetrans} & \underline{0.7976} & 0.3944 & \underline{0.6305} & \underline{0.6914} & 0.2328 & 71.89 & 0.7915 & 0.7851 \\ 
   StoryDiffusion \cite{zhou2024storydiffusion} & - & 0.5222 & 0.5806 & 0.6742 & 0.1489 & 92.21 & 0.4516 & 0.5873 \\
   Phantom \cite{liu2025phantom} & - & \underline{0.6211} & 0.6183 & 0.6793 & 0.2370 & 86.61 & 0.5379 & 0.5709 \\
   HunyuanVideo \cite{kong2024hunyuanvideo} & 0.4698 & 0.3222 & 0.6101 & 0.6158 & 0.2351 & 69.88 & 0.5703 & 0.6601 \\
   Wan2.2 \cite{wan2025wan} & 0.2165 & 0.1022 & 0.5885 & 0.6199 & \underline{0.2387} & \underline{69.48} & 0.6895 & 0.7547 \\
   SynCamMaster \cite{baisyncammaster} & - & 0.3033 & 0.5453 & 0.6177 & 0.1882 & 72.47 & \textbf{0.7949} & \textbf{0.8418} \\
   ReCamMaster \cite{bai2025recammaster} & 0.0266 & 0.0333 & 0.5493 & 0.6111 & 0.2320 & 71.51 & - & - \\
      \hline
   \methodname (\textbf{Ours}) & \textbf{0.8956} & \textbf{0.6744} & \textbf{0.6374} & \textbf{0.6984} & \textbf{0.2394} & \textbf{68.45} & \underline{0.7918} & \underline{0.8251} \\
       \bottomrule
  \end{tabular}%
\end{table*}

\begin{table*}[tb]
  \scriptsize
  \caption{Ablation results for shot-aware mask and training. The best is in \textbf{bold}.}
  \label{tab:ablation1}
  \centering
  \resizebox{\textwidth}{!}{%
  \renewcommand{\arraystretch}{1.2}%
  \begin{tabular}{l|cc|cccc|cc} 
    \toprule
    \multirow{2}{*}{Method} & \multicolumn{2}{c}{ \textbf{Transition Control}} & \multicolumn{4}{c}{\textbf{Overall Quality}} & \multicolumn{2}{c}{\textbf{Cross-shot Consistency}} \\ \cline{2-9}

    & Confidence\textuparrow & Type Acc\textuparrow & Aesthetic\textuparrow & Imaging\textuparrow & Overall Consistency\textuparrow & FVD\textdownarrow & Semantic\textuparrow & Visual\textuparrow \\ \hline
    
   \methodname (w/o Shot-aware Mask) & 0.7572 & 0.5422 & 0.6303 & 0.6912 & 0.2348 & 70.36 & 0.7183 & 0.7910 \\
   \methodname (w/o Semantic Mask) & 0.8913 & 0.6428 & 0.6332 & 0.6899 & 0.2371 & 71.54 & 0.6901 & 0.7761 \\
   \methodname (w/o Visual Mask) & 0.8044 & 0.5583 & 0.6305 & 0.6885 & 0.2351 & 69.47 & 0.7909 & 0.8052 \\ \hline
   \methodname (w/o Training) & 0.1402 & 0.2489 & 0.6276 & 0.6742 & 0.2233 & 70.71 & \textbf{0.8419} & \textbf{0.8256} \\ 
   \methodname (w/o Stage II Training) & 0.8615 & 0.6300 & 0.6331 & 0.6922 & 0.2379 & 68.97 & 0.7713 & 0.8076 \\
      \hline
   \methodname (\textbf{Ours}) & \textbf{0.8956} & \textbf{0.6744} & \textbf{0.6374} & \textbf{0.6984} & \textbf{0.2394} & \textbf{68.45} & 0.7918 & 0.8251 \\
       \bottomrule
  \end{tabular}%
  }
\end{table*}

\begin{table}[tb]
  \scriptsize
  \caption{Ablation results for camera information. The best is \textbf{bold}.}
  \label{tab:ablation2}
  \centering
  \begin{tabular}{l|cc} 
    \toprule
    Method & RotErr\textdownarrow & TransErr\textdownarrow \\
    \hline
    \methodname (w/o Camera Info) & 0.6330 & 0.5740 \\
    \methodname (w/o Plücker Branch) & 0.6262 & 0.5727 \\
    \methodname (w/o Extrinsic Branch) & 0.5972 & 0.5445 \\
    \hline
    \methodname (\textbf{Ours}) & \textbf{0.5907} & \textbf{0.5393} \\
       \bottomrule
  \end{tabular}%
\end{table}

\subsection{Evaluation}
\label{sec:evaluation}

\noindent \textbf{Evaluation Metrics}.
To comprehensively evaluate the performance of our model, we design a suite of evaluation prompts and metrics.   
The evaluation dataset consists of 90 prompts featuring hierarchical text captions and camera poses. 
We assess the generated results from three perspectives: shot transition control, overall quality, and cross-shot consistency. 
For transition control, we use TransNetV2 \cite{soucek2024transnet} to compute the Transition Confidence Score, evaluating the clarity of generated shot transitions. Qwen \cite{wang2024qwen2} is further employed to assess the Transition Type Accuracy, measuring whether the generated transition type aligns with the prompt.  
For overall quality, we adopt an aesthetic predictor  and an imaging quality model \cite{ke2021musiq} for visual assessment, and use ViCLIP \cite{internvid} feature similarity to quantify text-video alignment. Fréchet Video Distance \cite{unterthiner2018towards,unterthiner2019fvd} measures the distributional gap between generated and real film-edited videos.  
For consistency, we evaluate semantic and visual coherence across shots. Semantic consistency is measured via ViCLIP features extracted from each shot, while visual consistency is computed as the averaged subject \cite{oquab2023dinov2} and background \cite{fu2023dreamsim} similarity between adjacent shots.  
Together, this evaluation suite provides a scalable and systematic framework for assessing the quality and controllability of multi-shot video generation.

\begin{figure}[t]
  \centering
   \includegraphics[width=1.0\linewidth]{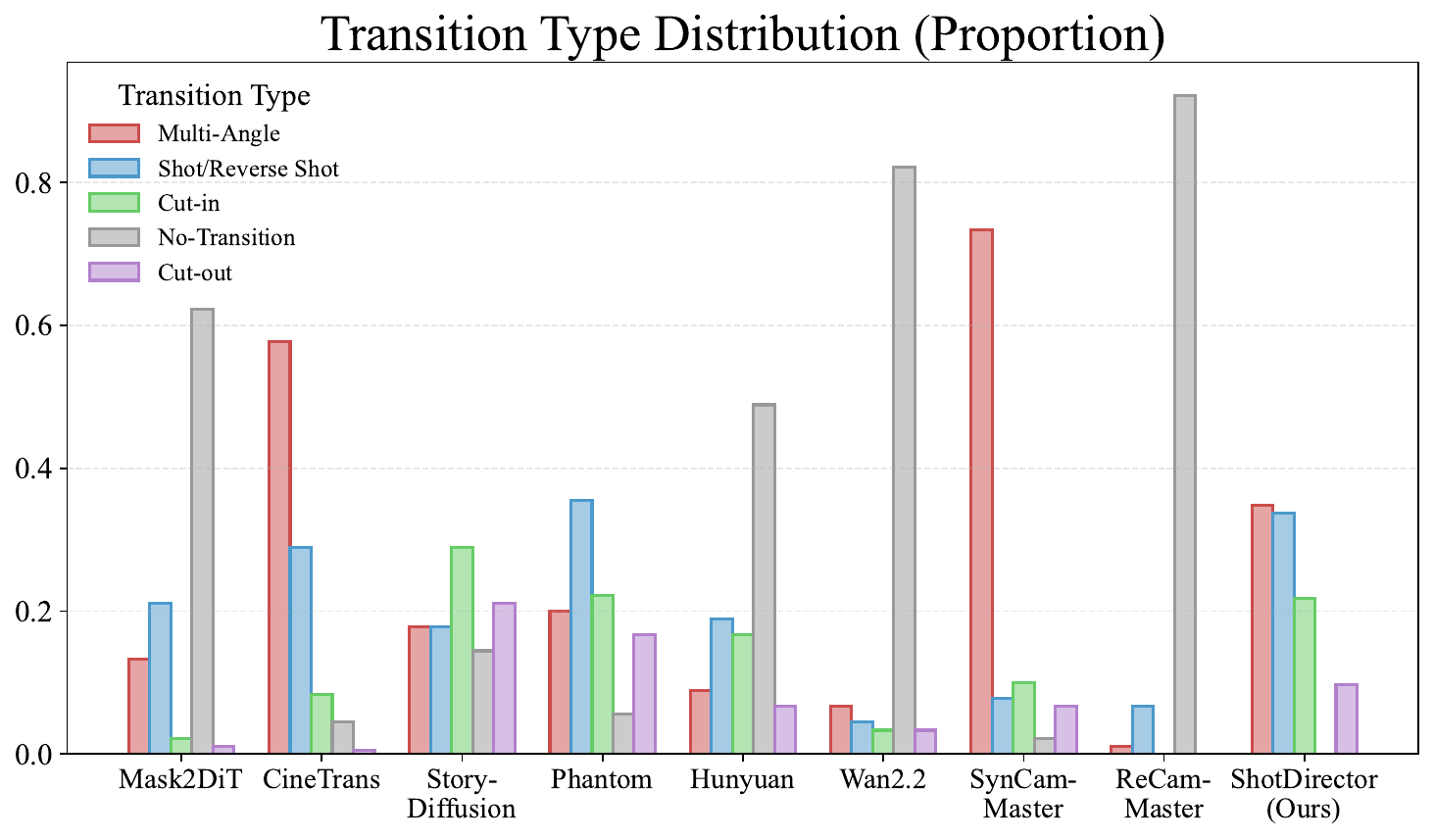}

   \caption{The distribution of shot transition types in videos generated by different methods. Our model (\methodname) stably produces transitions and achieves a more balanced distribution across transition types compared to other methods.}
   \label{fig:transitiontype}
\end{figure}

\noindent \textbf{Qualitative Results}.  
\cref{fig:results} presents a visual comparison between our method and representative baselines.  
For multi-shot video generation methods, Mask$^2$DiT exhibits instability when generating multiple shots and tends to produce animation-like visuals. CineTrans lacks a clear understanding of shot transition types, while StoryDiffusion and Phantom maintain moderate subject and style consistency but fail to preserve coherent visual details or form a continuous narrative flow.  
Among pre-trained video generation models, HunyuanVideo performs better than Wan 2.2 in multi-shot scenarios. Trained on large-scale datasets, both models demonstrate certain awareness of professional editing patterns. However, they cannot ensure the occurrence of multi-shot structures or explicitly control shot transitions.
For multi-view and camera-control methods, SynCamMaster maintains camera positioning and scene consistency but has no concept of shot transition type and produces relatively low-quality visuals. ReCamMaster, designed for smoothly varying camera poses, struggles with abrupt pose changes, resulting in distorted frames and failure to achieve shot transitions.  
In contrast, our method effectively responds to specified shot transition types and demonstrates film-like, professional editing patterns that convey coherent visual storytelling and semantic expression.

\noindent \textbf{Quantitative Results}. 
We conduct quantitative evaluations across the designed set of metrics, with results summarized in \cref{tab:evaluation}.  
As shown, most multi-shot video generation and pre-trained video generation models exhibit weak control over shot transitions. In contrast, \methodname achieves the most accurate transition control.  
In terms of overall quality, \methodname produces videos that are semantically aligned, visually high-quality, and closest to real multi-shot videos.  
Regarding consistency, although SynCamMaster achieves the highest scores, its strong camera pose constraints lead to low aesthetic quality and poor semantic adherence, indicating that its consistency is achieved at the cost of visual fidelity.  
Meanwhile, our model ranks second in consistency while maintaining superior overall quality, demonstrating that \methodname achieves both visual fidelity and temporal coherence.
We also visualize the distribution of shot transition types in videos generated by different methods in \cref{fig:transitiontype}. Some methods \cite{qi2025mask2dit,kong2024hunyuanvideo,wan2025wan,bai2025recammaster} exhibit limited transition capability, resulting in most videos being classified as No-Transition, while others \cite{wu2025cinetrans,baisyncammaster} lack explicit shot transition design and are predominantly labeled as Multi-Angle. In contrast, our method consistently produces diverse and professional shot transitions.

\begin{figure}[t]
  \centering
   \includegraphics[width=1.0\linewidth]{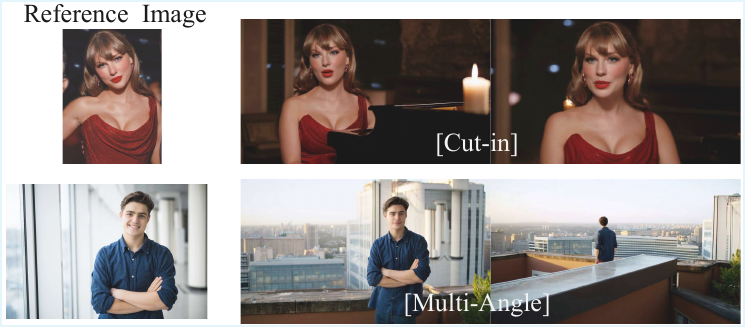}
   \caption{The performance of our method after being transferred to the reference-to-video model.}
   \label{fig:VACE}
\end{figure}

\subsection{Ablation Studies}
\label{sec:ablation}
In this section, we conduct ablation studies to evaluate the contribution of each component in our framework. 

\noindent \textbf{Camera Information Injection}. 
We compare variants using different camera information branches, as well as a version without camera conditioning. Following \cite{he2024cameractrl}, we adopt RotErr and TransErr to measure camera pose control performance (see \cref{tab:ablation2}).  
Results show that both camera information branches positively contribute to the capability of multi-shot video generation, with the Plücker branch performing slightly better than the extrinsic branch. This can be attributed to the intrinsic parameters and spatial ray maps in the Plücker representation, which help the model better interpret camera pose variations in shot transitions.

\noindent \textbf{Shot-Aware Mask Mechanism}.
We further ablate the visual and semantic shot-aware masks, as summarized in \cref{tab:ablation1}. The visual shot-aware mask has a stronger effect on transition control, as globally visible visual tokens may cause information leakage across shots, thereby reducing shot diversity and weakening transition effects.  
In contrast, the semantic shot-aware mask primarily influences consistency. This aligns with our intuition that fine-grained semantic control helps the model balance consistency and diversity in multi-shot scenarios.

\noindent \textbf{Training Strategy}.
Lastly, we perform ablations on the two-stage training process (\cref{tab:ablation1}). The results verify the effectiveness of our training strategy in enhancing both transition controllability and overall visual quality. 
Notably, the untrained version shows a higher consistency score, as it lacks the concept of multi-shot composition and fails to generate large visual variations in shot transitions, which is consistent with its lower transition confidence score.

\subsection{Additional Capability} 
\label{sec:addition}

As an extension, our model can seamlessly integrate functional modules trained on the base model to enable additional capabilities in multi-shot video generation, such as reference-to-video synthesis. As shown in \cref{fig:VACE}, we directly transfer the weights of \cite{vace} to ShotDirector, using reference images as additional inputs to generate multi-shot videos featuring specified subjects. 
This observation indicates that our model preserves the base model’s understanding of video content, allowing it to interface with other functional modules. Such adaptability further highlights the generalizability of our approach.


\section{Conclusion}
In this work, we presented \methodname, a unified framework for controllable multi-shot video generation that integrates parameter-level camera control with hierarchical, editing-pattern-aware prompting. By introducing explicit camera information injection and a shot-aware mask mechanism, our method enables fine-grained control over shot transitions while preserving semantic and visual consistency, effectively capturing film-like editing patterns and achieving  coherent narrative expression.

Our work highlights the importance of shot transition, including transition type, narrative flow, and cinematographic language, in multi-shot video generation task. Future research could extend this direction by exploring longer video sequences and more diverse, complex transition types, which would further pave the way for diffusion-based models to serve as directorial tools.

{
    \small
    \bibliographystyle{ieeenat_fullname}
    \bibliography{main}

@String(ICCV= {Int. Conf. Comput. Vis.})

@String(ICLR = {Int. Conf. Learn. Represent.})

@String(ICCV  = {ICCV})

@String(ICLR  = {ICLR})

@article{stablevideodiffusion,
  title={Stable video diffusion: Scaling latent video diffusion models to large datasets},
  author={Blattmann, Andreas and Dockhorn, Tim and Kulal, Sumith and Mendelevitch, Daniel and Kilian, Maciej and Lorenz, Dominik and Levi, Yam and English, Zion and Voleti, Vikram and Letts, Adam and others},
  journal={arXiv preprint arXiv:2311.15127},
  year={2023}
}

@inproceedings{ho2020denoising,
  title={Denoising diffusion probabilistic models},
  author={Ho, Jonathan and Jain, Ajay and Abbeel, Pieter},
  booktitle={Neural Information Processing Systems},
  volume={33},
  pages={6840--6851},
  year={2020}
}

@inproceedings{rombach2022high,
  title={High-resolution image synthesis with latent diffusion models},
  author={Rombach, Robin and Blattmann, Andreas and Lorenz, Dominik and Esser, Patrick and Ommer, Bj{\"o}rn},
  booktitle={Computer Vision and Pattern Recognition},
  pages={10684--10695},
  year={2022}
}

@inproceedings{song2020score,
  title={Score-Based Generative Modeling through Stochastic Differential Equations},
  author={Song, Yang and Sohl-Dickstein, Jascha and Kingma, Diederik P and Kumar, Abhishek and Ermon, Stefano and Poole, Ben},
  booktitle= {International Conference on Learning Representations}
}

@article{wan2025wan,
  title={Wan: Open and advanced large-scale video generative models},
  author={Wan, Team and Wang, Ang and Ai, Baole and Wen, Bin and Mao, Chaojie and Xie, Chen-Wei and Chen, Di and Yu, Feiwu and Zhao, Haiming and Yang, Jianxiao and others},
  journal={arXiv preprint arXiv:2503.20314},
  year={2025}
}

@article{kong2024hunyuanvideo,
  title={Hunyuanvideo: A systematic framework for large video generative models},
  author={Kong, Weijie and Tian, Qi and Zhang, Zijian and Min, Rox and Dai, Zuozhuo and Zhou, Jin and Xiong, Jiangfeng and Li, Xin and Wu, Bo and Zhang, Jianwei and others},
  journal={arXiv preprint arXiv:2412.03603},
  year={2024}
}

@article{li2024longvideosurvey,
  title={A survey on long video generation: Challenges, methods, and prospects},
  author={Li, Chengxuan and Huang, Di and Lu, Zeyu and Xiao, Yang and Pei, Qingqi and Bai, Lei},
  journal={arXiv preprint arXiv:2403.16407},
  year={2024}
}

@article{wang2024lavie,
  title={Lavie: High-quality video generation with cascaded latent diffusion models},
  author={Wang, Yaohui and Chen, Xinyuan and Ma, Xin and Zhou, Shangchen and Huang, Ziqi and Wang, Yi and Yang, Ceyuan and He, Yinan and Yu, Jiashuo and Yang, Peiqing and others},
  journal={International Journal of Computer Vision},
  pages={1--20},
  year={2024},
  publisher={Springer}
}

@inproceedings{DiT,
  title={Scalable diffusion models with transformers},
  author={Peebles, William and Xie, Saining},
  booktitle={Proceedings of the IEEE/CVF international conference on computer vision},
  pages={4195--4205},
  year={2023}
}

@article{sora,
  title={Video generation models as world simulators},
  author={Brooks, Tim and Peebles, Bill and Holmes, Connor and DePue, Will and Guo, Yufei and Jing, Li and Schnurr, David and Taylor, Joe and Luhman, Troy and Luhman, Eric and others},
  journal={OpenAI Blog},
  volume={1},
  number={8},
  pages={1},
  year={2024}
}

@article{yang2024cogvideox,
  title={Cogvideox: Text-to-video diffusion models with an expert transformer},
  author={Yang, Zhuoyi and Teng, Jiayan and Zheng, Wendi and Ding, Ming and Huang, Shiyu and Xu, Jiazheng and Yang, Yuanming and Hong, Wenyi and Zhang, Xiaohan and Feng, Guanyu and others},
  journal={arXiv preprint arXiv:2408.06072},
  year={2024}
}

@inproceedings{qi2025mask2dit,
  title={Mask\^{} 2DiT: Dual Mask-based Diffusion Transformer for Multi-Scene Long Video Generation},
  author={Qi, Tianhao and Yuan, Jianlong and Feng, Wanquan and Fang, Shancheng and Liu, Jiawei and Zhou, SiYu and He, Qian and Xie, Hongtao and Zhang, Yongdong},
  booktitle={Proceedings of the Computer Vision and Pattern Recognition Conference},
  pages={18837--18846},
  year={2025}
}

@InProceedings{guo2025longcontexttuning,
    author    = {Guo, Yuwei and Yang, Ceyuan and Yang, Ziyan and Ma, Zhibei and Lin, Zhijie and Yang, Zhenheng and Lin, Dahua and Jiang, Lu},
    title     = {Long Context Tuning for Video Generation},
    booktitle = {Proceedings of the IEEE/CVF International Conference on Computer Vision (ICCV)},
    month     = {October},
    year      = {2025},
    pages     = {17281-17291}
}

@article{wu2025cinetrans,
  title={CineTrans: Learning to Generate Videos with Cinematic Transitions via Masked Diffusion Models},
  author={Wu, Xiaoxue and Gao, Bingjie and Qiao, Yu and Wang, Yaohui and Chen, Xinyuan},
  journal={arXiv preprint arXiv:2508.11484},
  year={2025}
}

@inproceedings{ttt,
  title={One-minute video generation with test-time training},
  author={Dalal, Karan and Koceja, Daniel and Xu, Jiarui and Zhao, Yue and Han, Shihao and Cheung, Ka Chun and Kautz, Jan and Choi, Yejin and Sun, Yu and Wang, Xiaolong},
  booktitle={Proceedings of the Computer Vision and Pattern Recognition Conference},
  pages={17702--17711},
  year={2025}
}

@article{zhou2024storydiffusion,
  title={Storydiffusion: Consistent self-attention for long-range image and video generation},
  author={Zhou, Yupeng and Zhou, Daquan and Cheng, Ming-Ming and Feng, Jiashi and Hou, Qibin},
  journal={Advances in Neural Information Processing Systems},
  volume={37},
  pages={110315--110340},
  year={2024}
}

@article{zheng2024videogenofthought,
  title={VideoGen-of-Thought: A Collaborative Framework for Multi-Shot Video Generation},
  author={Zheng, Mingzhe and Xu, Yongqi and Huang, Haojian and Ma, Xuran and Liu, Yexin and Shu, Wenjie and Pang, Yatian and Tang, Feilong and Chen, Qifeng and Yang, Harry and Lim Sernam},
  journal={arXiv preprint arXiv:2412.02259},
  year={2024}
}

@inproceedings{long2024videostudio,
  title={Videostudio: Generating consistent-content and multi-scene videos},
  author={Long, Fuchen and Qiu, Zhaofan and Yao, Ting and Mei, Tao},
  booktitle={European Conference on Computer Vision},
  pages={468--485},
  year={2024},
  organization={Springer}
}

@article{he2023animateastory,
  title={Animate-a-story: Storytelling with retrieval-augmented video generation},
  author={He, Yingqing and Xia, Menghan and Chen, Haoxin and Cun, Xiaodong and Gong, Yuan and Xing, Jinbo and Zhang, Yong and Wang, Xintao and Weng, Chao and Shan, Ying and others},
  journal={arXiv preprint arXiv:2307.06940},
  year={2023}
}

@article{xie2024dreamfactory,
  title={Dreamfactory: Pioneering multi-scene long video generation with a multi-agent framework},
  author={Xie, Zhifei and Tang, Daniel and Tan, Dingwei and Klein, Jacques and Bissyand, Tegawend F and Ezzini, Saad},
  journal={arXiv preprint arXiv:2408.11788},
  year={2024}
}

@article{zhao2024moviedreamer,
  title={Moviedreamer: Hierarchical generation for coherent long visual sequence},
  author={Zhao, Canyu and Liu, Mingyu and Wang, Wen and Chen, Weihua and Wang, Fan and Chen, Hao and Zhang, Bo and Shen, Chunhua},
  journal={arXiv preprint arXiv:2407.16655},
  year={2024}
}

@article{jia2025moga,
  title={MoGA: Mixture-of-Groups Attention for End-to-End Long Video Generation},
  author={Jia, Weinan and Lu, Yuning and Huang, Mengqi and Wang, Hualiang and Huang, Binyuan and Chen, Nan and Liu, Mu and Jiang, Jidong and Mao, Zhendong},
  journal={arXiv preprint arXiv:2510.18692},
  year={2025}
}

@article{sitzmann2021light,
  title={Light field networks: Neural scene representations with single-evaluation rendering},
  author={Sitzmann, Vincent and Rezchikov, Semon and Freeman, Bill and Tenenbaum, Josh and Durand, Fredo},
  journal={Advances in Neural Information Processing Systems},
  volume={34},
  pages={19313--19325},
  year={2021}
}

@article{xing2024survey,
  title={A survey on video diffusion models},
  author={Xing, Zhen and Feng, Qijun and Chen, Haoran and Dai, Qi and Hu, Han and Xu, Hang and Wu, Zuxuan and Jiang, Yu-Gang},
  journal={ACM Computing Surveys},
  volume={57},
  number={2},
  pages={1--42},
  year={2024},
  publisher={ACM New York, NY}
}

@inproceedings{khachatryan2023text2video,
  title={Text2video-zero: Text-to-image diffusion models are zero-shot video generators},
  author={Khachatryan, Levon and Movsisyan, Andranik and Tadevosyan, Vahram and Henschel, Roberto and Wang, Zhangyang and Navasardyan, Shant and Shi, Humphrey},
  booktitle={Proceedings of the IEEE/CVF International Conference on Computer Vision},
  pages={15954--15964},
  year={2023}
}

@inproceedings{zhuang2024vlogger,
  title={Vlogger: Make your dream a vlog},
  author={Zhuang, Shaobin and Li, Kunchang and Chen, Xinyuan and Wang, Yaohui and Liu, Ziwei and Qiao, Yu and Wang, Yali},
  booktitle={Proceedings of the IEEE/CVF Conference on Computer Vision and Pattern Recognition},
  pages={8806--8817},
  year={2024}
}

@article{ma2024latte,
  title={Latte: Latent diffusion transformer for video generation},
  author={Ma, Xin and Wang, Yaohui and Jia, Gengyun and Chen, Xinyuan and Liu, Ziwei and Li, Yuan-Fang and Chen, Cunjian and Qiao, Yu},
  journal={arXiv preprint arXiv:2401.03048},
  year={2024}
}

@article{liu2025phantom,
  title={Phantom: Subject-consistent video generation via cross-modal alignment},
  author={Liu, Lijie and Ma, Tianxiang and Li, Bingchuan and Chen, Zhuowei and Liu, Jiawei and Li, Gen and Zhou, Siyu and He, Qian and Wu, Xinglong},
  journal={arXiv preprint arXiv:2502.11079},
  year={2025}
}

@article{hu2022lora,
  title={Lora: Low-rank adaptation of large language models.},
  author={Hu, Edward J and Shen, Yelong and Wallis, Phillip and Allen-Zhu, Zeyuan and Li, Yuanzhi and Wang, Shean and Wang, Lu and Chen, Weizhu and others},
  journal={ICLR},
  volume={1},
  number={2},
  pages={3},
  year={2022}
}

@inproceedings{kara2025shotadapter,
  title={ShotAdapter: Text-to-Multi-Shot Video Generation with Diffusion Models},
  author={Kara, Ozgur and Singh, Krishna Kumar and Liu, Feng and Ceylan, Duygu and Rehg, James M and Hinz, Tobias},
  booktitle={Proceedings of the Computer Vision and Pattern Recognition Conference},
  pages={28405--28415},
  year={2025}
}

@article{he2025cut2next,
  title={Cut2Next: Generating Next Shot via In-Context Tuning},
  author={He, Jingwen and Liu, Hongbo and Li, Jiajun and Huang, Ziqi and Qiao, Yu and Ouyang, Wanli and Liu, Ziwei},
  journal={arXiv preprint arXiv:2508.08244},
  year={2025}
}

@article{he2024cameractrl,
      title={CameraCtrl: Enabling Camera Control for Text-to-Video Generation}, 
      author={Hao He and Yinghao Xu and Yuwei Guo and Gordon Wetzstein and Bo Dai and Hongsheng Li and Ceyuan Yang},
      journal={arXiv preprint arXiv:2404.02101},
      year={2024}
}

@inproceedings{fengi2vcontrol,
  title={I2VControl-Camera: Precise Video Camera Control with Adjustable Motion Strength},
  author={Feng, Wanquan and Liu, Jiawei and Tu, Pengqi and Qi, Tianhao and Sun, Mingzhen and Ma, Tianxiang and Zhao, Songtao and Zhou, SiYu and HE, Qian},
  booktitle={The Thirteenth International Conference on Learning Representations},
  year={2025}
}

@inproceedings{bahmani2025ac3d,
  title={Ac3d: Analyzing and improving 3d camera control in video diffusion transformers},
  author={Bahmani, Sherwin and Skorokhodov, Ivan and Qian, Guocheng and Siarohin, Aliaksandr and Menapace, Willi and Tagliasacchi, Andrea and Lindell, David B and Tulyakov, Sergey},
  booktitle={Proceedings of the Computer Vision and Pattern Recognition Conference},
  pages={22875--22889},
  year={2025}
}

@inproceedings{bahmanivd3d,
  title={VD3D: Taming Large Video Diffusion Transformers for 3D Camera Control},
  author={Bahmani, Sherwin and Skorokhodov, Ivan and Siarohin, Aliaksandr and Menapace, Willi and Qian, Guocheng and Vasilkovsky, Michael and Lee, Hsin-Ying and Wang, Chaoyang and Zou, Jiaxu and Tagliasacchi, Andrea and others},
  booktitle={The Thirteenth International Conference on Learning Representations},
  year={2025}
}

@inproceedings{houtraining,
  title={Training-free Camera Control for Video Generation},
  author={Hou, Chen and Chen, Zhibo},
  booktitle={The Thirteenth International Conference on Learning Representations},
  year={2025}
}

@article{xu2024camco,
  title={Camco: Camera-controllable 3d-consistent image-to-video generation},
  author={Xu, Dejia and Nie, Weili and Liu, Chao and Liu, Sifei and Kautz, Jan and Wang, Zhangyang and Vahdat, Arash},
  journal={arXiv preprint arXiv:2406.02509},
  year={2024}
}

@inproceedings{baisyncammaster,
  title={SynCamMaster: Synchronizing Multi-Camera Video Generation from Diverse Viewpoints},
  author={Bai, Jianhong and Xia, Menghan and Wang, Xintao and Yuan, Ziyang and Liu, Zuozhu and Hu, Haoji and Wan, Pengfei and ZHANG, Di},
  booktitle={The Thirteenth International Conference on Learning Representations},
  year={2025}
}

@article{bai2025recammaster,
  title={Recammaster: Camera-controlled generative rendering from a single video},
  author={Bai, Jianhong and Xia, Menghan and Fu, Xiao and Wang, Xintao and Mu, Lianrui and Cao, Jinwen and Liu, Zuozhu and Hu, Haoji and Bai, Xiang and Wan, Pengfei and others},
  journal={arXiv preprint arXiv:2503.11647},
  year={2025}
}

@article{he2025cameractrl,
  title={Cameractrl ii: Dynamic scene exploration via camera-controlled video diffusion models},
  author={He, Hao and Yang, Ceyuan and Lin, Shanchuan and Xu, Yinghao and Wei, Meng and Gui, Liangke and Zhao, Qi and Wetzstein, Gordon and Jiang, Lu and Li, Hongsheng},
  journal={arXiv preprint arXiv:2503.10592},
  year={2025}
}

@inproceedings{wang2025vggt,
  title={Vggt: Visual geometry grounded transformer},
  author={Wang, Jianyuan and Chen, Minghao and Karaev, Nikita and Vedaldi, Andrea and Rupprecht, Christian and Novotny, David},
  booktitle={Proceedings of the Computer Vision and Pattern Recognition Conference},
  pages={5294--5306},
  year={2025}
}

@inproceedings{soucek2024transnet,
  title={Transnet v2: An effective deep network architecture for fast shot transition detection},
  author={Soucek, Tom{\'a}s and Lokoc, Jakub},
  booktitle={Proceedings of the 32nd ACM International Conference on Multimedia},
  pages={11218--11221},
  year={2024}
}

@inproceedings{girdhar2023imagebind,
  title={Imagebind: One embedding space to bind them all},
  author={Girdhar, Rohit and El-Nouby, Alaaeldin and Liu, Zhuang and Singh, Mannat and Alwala, Kalyan Vasudev and Joulin, Armand and Misra, Ishan},
  booktitle={Proceedings of the IEEE/CVF conference on computer vision and pattern recognition},
  pages={15180--15190},
  year={2023}
}

@inproceedings{radford2021learning,
  title={Learning transferable visual models from natural language supervision},
  author={Radford, Alec and Kim, Jong Wook and Hallacy, Chris and Ramesh, Aditya and Goh, Gabriel and Agarwal, Sandhini and Sastry, Girish and Askell, Amanda and Mishkin, Pamela and Clark, Jack and others},
  booktitle={International conference on machine learning},
  pages={8748--8763},
  year={2021},
  organization={PmLR}
}

@article{wang2024qwen2,
  title={Qwen2-vl: Enhancing vision-language model's perception of the world at any resolution},
  author={Wang, Peng and Bai, Shuai and Tan, Sinan and Wang, Shijie and Fan, Zhihao and Bai, Jinze and Chen, Keqin and Liu, Xuejing and Wang, Jialin and Ge, Wenbin and others},
  journal={arXiv preprint arXiv:2409.12191},
  year={2024}
}

@article{adam2014method,
  title={A method for stochastic optimization},
  author={Adam, Kingma DP Ba J and others},
  journal={arXiv preprint arXiv:1412.6980},
  volume={1412},
  number={6},
  year={2014}
}

@inproceedings{ke2021musiq,
  title={Musiq: Multi-scale image quality transformer},
  author={Ke, Junjie and Wang, Qifei and Wang, Yilin and Milanfar, Peyman and Yang, Feng},
  booktitle={Proceedings of the IEEE/CVF international conference on computer vision},
  pages={5148--5157},
  year={2021}
}

@article{internvid,
  title={Internvid: A large-scale video-text dataset for multimodal understanding and generation},
  author={Wang, Yi and He, Yinan and Li, Yizhuo and Li, Kunchang and Yu, Jiashuo and Ma, Xin and Li, Xinhao and Chen, Guo and Chen, Xinyuan and Wang, Yaohui and others},
  journal={arXiv preprint arXiv:2307.06942},
  year={2023}
}

@inproceedings{unterthiner2019fvd,
  title={FVD: A new metric for video generation},
  author={Unterthiner, Thomas and Van Steenkiste, Sjoerd and Kurach, Karol and Marinier, Rapha{\"e}l and Michalski, Marcin and Gelly, Sylvain},
  booktitle={ICLRW},
  year={2019}
}

@article{unterthiner2018towards,
  title={Towards accurate generative models of video: A new metric \& challenges},
  author={Unterthiner, Thomas and Van Steenkiste, Sjoerd and Kurach, Karol and Marinier, Raphael and Michalski, Marcin and Gelly, Sylvain},
  journal={arXiv preprint arXiv:1812.01717},
  year={2018}
}

@article{oquab2023dinov2,
  title={DINOv2: Learning Robust Visual Features without Supervision},
  author={Oquab, Maxime and Darcet, Timoth{\'e}e and Moutakanni, Th{\'e}o and Vo, Huy and Szafraniec, Marc and Khalidov, Vasil and Fernandez, Pierre and Haziza, Daniel and Massa, Francisco and El-Nouby, Alaaeldin and others},
  journal={Transactions on Machine Learning Research Journal},
  pages={1--31},
  year={2024}
}

@inproceedings{fu2023dreamsim,
  title={DreamSim: Learning New Dimensions of Human Visual Similarity using Synthetic Data},
  author={Fu, Stephanie and Tamir, Netanel and Sundaram, Shobhita and Chai, Lucy and Zhang, Richard and Dekel, Tali and Isola, Phillip},
  booktitle={Neural Information Processing Systems},
  volume={36},
  pages={50742--50768},
  year={2023}
}

@inproceedings{vace,
    title = {VACE: All-in-One Video Creation and Editing},
    author = {Jiang, Zeyinzi and Han, Zhen and Mao, Chaojie and Zhang, Jingfeng and Pan, Yulin and Liu, Yu},
    booktitle = {Proceedings of the IEEE/CVF International Conference on Computer Vision},
    pages = {17191-17202},
    year = {2025}
}

@inproceedings{
chen2024seine,
title={{SEINE}: Short-to-Long Video Diffusion Model for Generative Transition and Prediction},
author={Xinyuan Chen and Yaohui Wang and Lingjun Zhang and Shaobin Zhuang and Xin Ma and Jiashuo Yu and Yali Wang and Dahua Lin and Yu Qiao and Ziwei Liu},
booktitle={The Twelfth International Conference on Learning Representations},
year={2024},
url={https://openreview.net/forum?id=FNq3nIvP4F}
}
}

\clearpage
\setcounter{page}{1}
\maketitlesupplementary
\appendix




\section{Details of Dataset Construction}

In \cref{sec:dataset}, we introduce \datasetname, a high-quality dataset with detailed shot transition annotations. This section provides additional details of the dataset construction pipeline to further support clarity and reproducibility.

\noindent \textbf{Raw Video Source.} 
To capture rich cinematography language and maintain narrative flow, we collect 16K full-length films as our raw video source. This large corpus provides shot transitions with strong semantic coherence.

\noindent \textbf{Segmentation and Stitching.} 
To obtain initial multi-shot clips that exhibit both semantic and visual similarity, we perform segmentation and similar-segment stitching on the raw videos. Specifically, we first segment each video using TransNetV2 \cite{soucek2024transnet}, and extract image features for adjacent shots using ImageBind \cite{girdhar2023imagebind}. A clip is discarded if the similarity between its first and last frames falls below a predefined threshold. The thresholds used during this stage are summarized in \cref{tab:seg}.

\begin{table}[h]
    \centering
    \caption{Threshold settings for the segmentation and stitching.}
    \label{tab:seg}
    \begin{tabular}{lc}
    \toprule
    \textbf{Threshold Type} & \textbf{Value} \\
    \hline
    Segmentation threshold & 0.45 \\
    First/last frame similarity threshold & 0.90 \\
    Stitching threshold & 0.65 \\
    \bottomrule
    \end{tabular}
\end{table}

\noindent \textbf{Coarse Filtering.}
After acquiring the initial multi-shot clips, we apply a coarse filtering stage based on fundamental video attributes, including frame rate, spatial resolution, temporal duration (5-12 seconds), and overall aesthetic quality.
To further ensure clear transition boundaries, the aesthetic score filtering is applied with particular attention to frames near the shot boundary, which helps prevent ambiguous or visually unclear transitions.
As our study focuses on shot transition types, which are defined between compositions of shot pair, we retain only clips containing two shots. This stage yields a pool of approximately 500K candidate videos.

\noindent \textbf{Fine-Grained Transition Filtering.}
Inter-shot consistency is crucial for selecting meaningful shot transitions. Extremely low similarity leads to abrupt or implausible visual break, while excessively high similarity reduces diversity and often corresponds to flash-like effects rather than genuine transitions. Although the segmentation-and-stitching stage provides coarse control, a more refined selection step is required to ensure spatial or causal coherence between shots and avoid potential confusion during model training.

\begin{figure*}[t]
  \centering
   \includegraphics[width=1.0\linewidth]{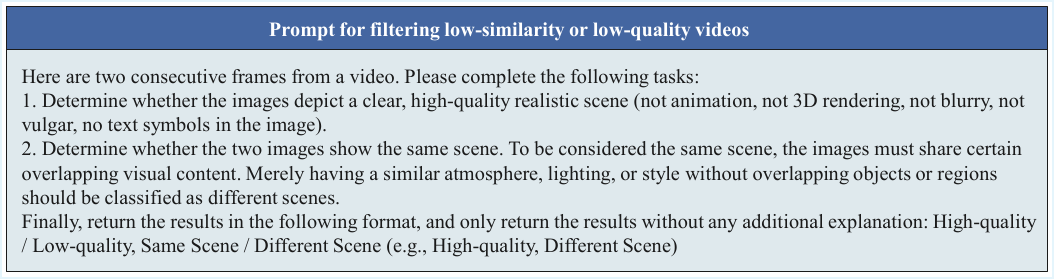}
   \caption{Prompt for filtering low-similarity or low-quality videos using Qwen.}
   \label{fig:Qwen}
\end{figure*}

To address excessive similarity, we compute CLIP \cite{radford2021learning} feature similarity and remove pairs with similarity greater than 0.95, effectively filtering out cases of incorrect segmentation or no-transition scenes (e.g., light flickering). For low similarity, we use a VLM-based \cite{wang2024qwen2} method, as image-feature-based metrics tend to focus on vibe, style, and tone rather than spatial or causal relationships. The prompts used for VLM filtering are shown in \cref{fig:Qwen}. This fine-grained filtering yields a final set of 40K videos.

\begin{figure*}[h]
  \centering
   \includegraphics[width=1.0\linewidth]{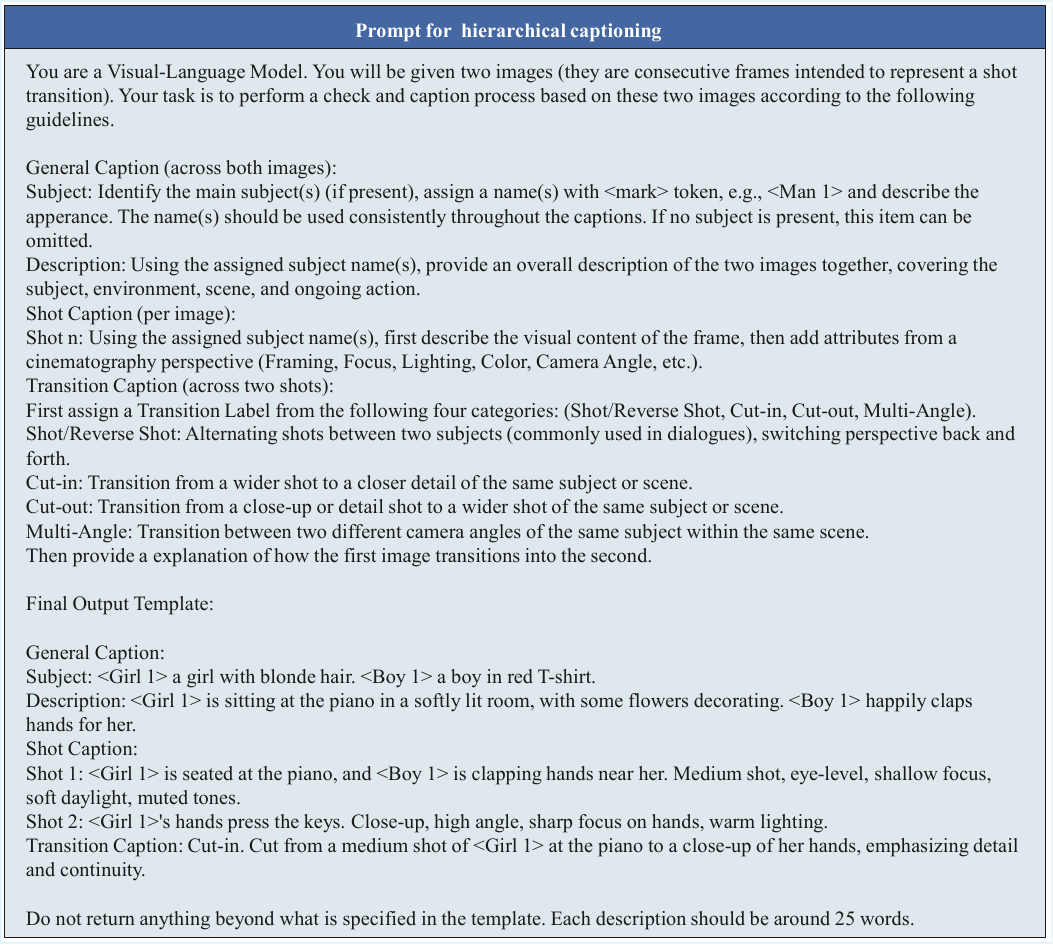}
   \caption{Prompt for hierarchical captioning using GPT-5-mini.}
   \label{fig:GPT}
\end{figure*}

\noindent \textbf{Caption Generation.}
We employ GPT-5-mini to generate hierarchical captions for each curated video. As shown in \cref{fig:GPT}, each video is annotated with a general description that captures the main subjects across the two shots, together with more fine-grained shot-specific captions that articulate the relevant cinematographic characteristics. This hierarchical annotation scheme endows the dataset with structured and professionally oriented cinematic transition knowledge.

\section{Statistic of Dataset}

\begin{figure*}[t]
  \centering
   \includegraphics[width=1.0\linewidth]{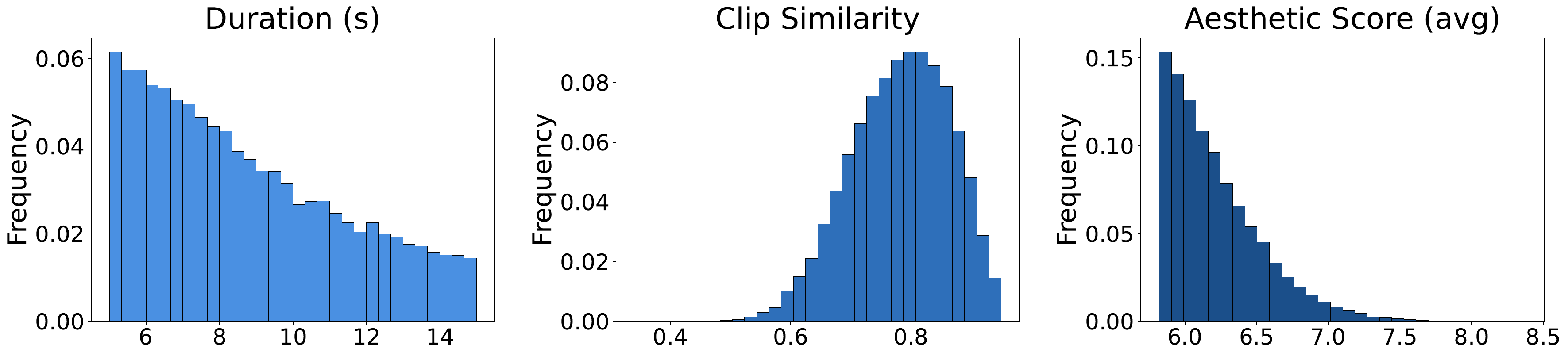}
   \caption{Distribution of key video attributes in \datasetname.}
   \label{fig:visualization}
\end{figure*}

This section presents detailed statistical characteristics of \datasetname, and visualizes the distributions of aesthetic score, shot duration, and inter-shot CLIP feature similarity in \cref{fig:visualization}.
Across the 40K curated two-shot clips, the dataset exhibits an average duration of 8.72 seconds, an average aesthetic score of 6.21, and a mean CLIP feature similarity of 0.7817 between adjacent-shot frame pairs.
These statistics indicate that the dataset maintains high aesthetic quality and strictly controlled inter-shot consistency, making it well suited for training frameworks aimed at the preliminary exploration of shot transition modeling.

\section{Complete Results of Qualitative Results}

\begin{figure*}[t]
  \centering
   \includegraphics[width=1.0\linewidth]{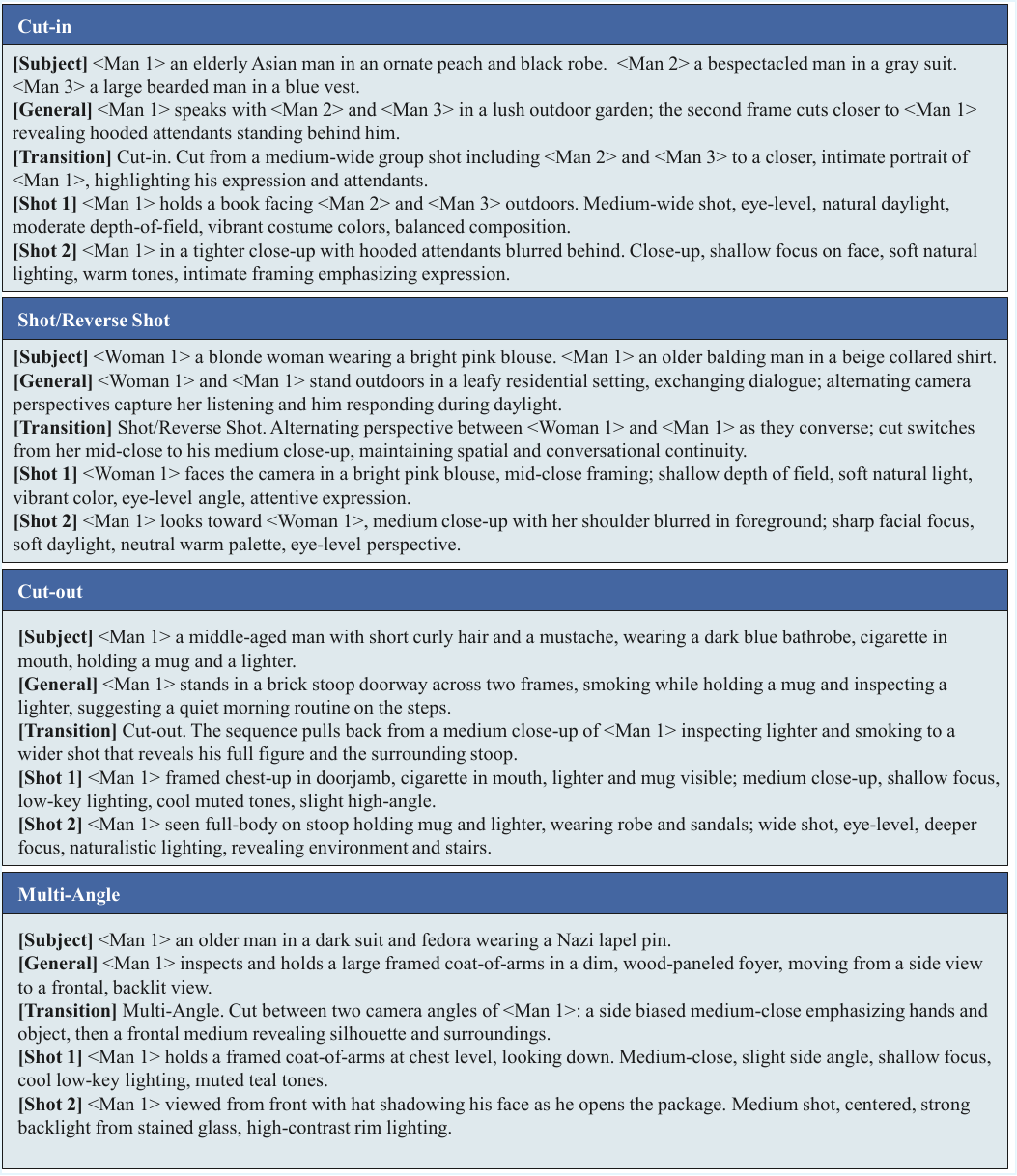}
   \caption{Prompt for qualitative evaluation using different models. (The results in video form are on the project page.)}
   \label{fig:four}
\end{figure*}

In \cref{sec:evaluation}, the transition-type-aware multi-shot video generation performance of \methodname is compared against several baseline models, and representative qualitative results are presented in \cref{fig:results}. This section provides the full set of qualitative visualizations, offering a more comprehensive and convincing analysis that further supports the effectiveness of \methodname.
We present the results of different models in video form on the project page provided in the supplementary material, and the corresponding full prompts are included in \cref{fig:four}. These extensive comparisons enable a more thorough examination of the strengths and limitations of each method.

For multi-shot video generation methods, shot-by-shot stitching approaches exhibit low consistency across shots. StoryDiffusion \cite{zhou2024storydiffusion} produces frames with substantial disparities between shots, preventing coherent multi-shot composition. Phantom \cite{liu2025phantom}, which incorporates reference images, improves subject consistency but struggles with background continuity, failing to maintain a unified scene and thus breaking the narrative flow.

For end-to-end multi-shot generation models, CineTrans \cite{wu2025cinetrans} is capable of producing shot transitions but lacks an understanding of transition-type semantics, resulting in outputs without professional cinematic characteristics. Mask2DiT \cite{qi2025mask2dit} tends to generate animation-like visuals, leading to relatively low perceptual quality.

Regarding large-scale pretrained models, HunyuanVideo \cite{kong2024hunyuanvideo} and Wan2.2 \cite{wan2025wan} demonstrate certain transition effects owing to their strong semantic understanding, yet the outcomes remain unstable, and neither consistency nor transition type can be reliably preserved.

For camera-controlled methods, SynCamMaster \cite{baisyncammaster} achieves reasonable camera control but at the expense of visual quality, likely due to the reliance on synthetic training data, which causes notable deviations from real-world video appearance. ReCamMaster \cite{bai2025recammaster} tends to perform camera motions instead of hard shot transitions and may suffer from visual artifacts during switching.

Across all these comparisons, the proposed framework demonstrates the ability to produce stable transitions while maintaining an understanding of professional cinematographic language, thereby enabling more controllable and semantically grounded shot transitions.
\section{Details of Evaluation Metrics}

In \cref{sec:evaluation}, a comprehensive evaluation protocol is introduced to quantitatively assess the performance of the proposed framework. This section provides additional details regarding the computation of several key metrics, further supporting the reproducibility and validity of the evaluation.

\noindent \textbf{Transition Confidence Score.}
To quantify the clarity and reliability of shot transitions in generated videos, the Transition Confidence Score is computed using TransNetV2 \cite{soucek2024transnet}.
Given an input video sequence, TransNetV2 produces a frame-wise transition likelihood feature $d\in \mathbb{R}^{F\times 1}$, where each element corresponds to the transition probability of a specific frame after applying a sigmoid activation.
The score for the entire video is defined as the maximum confidence over all frames:
\begin{equation}
\text{Transition Confidence Score} = \max (\sigma(d)),
\end{equation}
where $\sigma(\cdot)$ denotes the sigmoid function.
This metric captures whether a transition occurs within the sequence and how sharply it is presented (e.g., distinguishing hard cuts from gradual transitions), providing an intuitive measure of the model’s ability to generate well-structured shot transitions.

\begin{figure*}[t]
  \centering
   \includegraphics[width=1.0\linewidth]{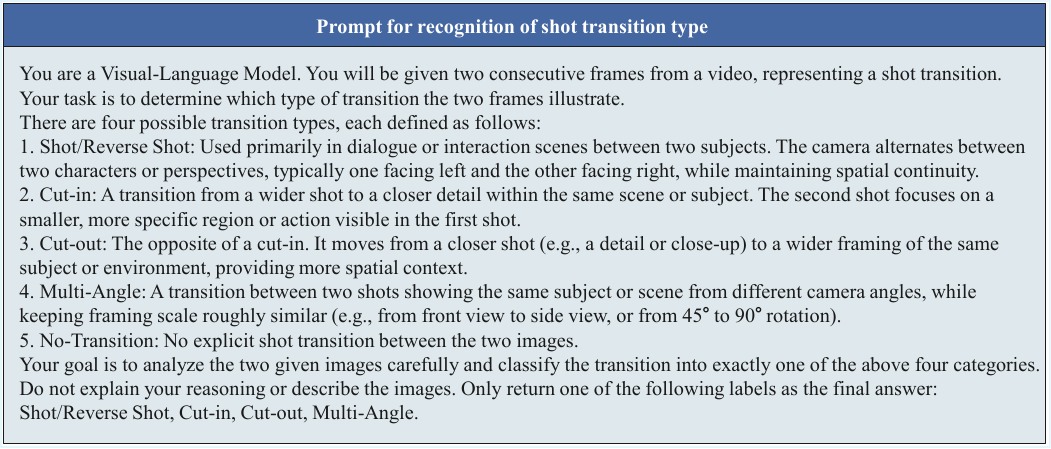}
   \caption{Prompt for recognition of shot transition type using Qwen.}
   \label{fig:transition-vlm-prompt}
\end{figure*}

\noindent \textbf{Transition Type Accuracy.}
Beyond detecting transitions, correctly modeling transition types is crucial for evaluating transition-aware multi-shot video generation.
To this end, Transition Type Accuracy is introduced to assess a model’s ability to adapt to different categories of transitions. A vision–language model (VLM) \cite{wang2024qwen2} is employed to classify the transition type of each generated video, and accuracy is computed against the ground-truth prompts.
\cref{tab:transition-stats} reports the distribution of transition types within the evaluation set, and Figure~\ref{fig:transition-vlm-prompt} illustrates the prompt used for VLM-based recognition.

\begin{table}[h]
    \centering
    \caption{Distribution of transition types in the evaluation set.}
    \label{tab:transition-stats}
    \begin{tabular}{lc}
    \toprule
    \textbf{Transition Type} & \textbf{Count} \\
    \midrule
    Cut-in            & 24 \\
    Cut-out           & 26 \\
    Shot/Reverse Shot & 25 \\
    Multi-Angle       & 15 \\
    \bottomrule
    \end{tabular}
\end{table}

\section{Limitation}

\subsection{Failure Case}

\begin{figure}[h]
  \centering
   \includegraphics[width=1.0\linewidth]{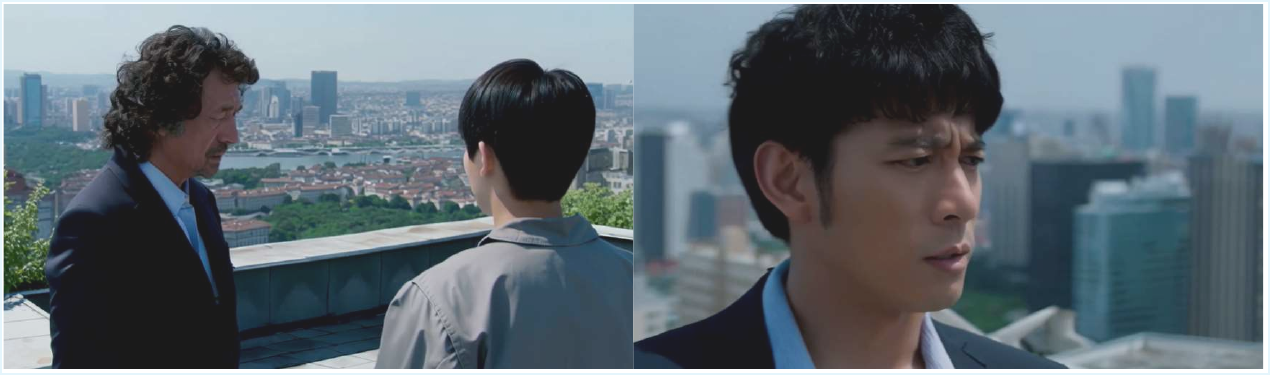}
   \caption{A representative failure case in which the visual characteristics of multiple subjects become unintentionally blended during generation.}
   \label{fig:fail}
\end{figure}

\cref{fig:fail} presents failure cases observed during inference. We find that in certain samples the visual characteristics of different subjects become mixed, suggesting that the model lacks a clear one-to-one correspondence when multiple subjects are present. This issue may indicate insufficient understanding of multi-subject scenarios within the model. A possible improvement is to provide more detailed bounding-box-level annotations in the dataset, which could enhance the model’s ability to better understand and model multi-subject situations.

\subsection{Future Work}

Although \methodname demonstrates strong performance in \cref{sec:evaluation}, it still exhibits several limitations and opens up promising avenues for future exploration.
\begin{itemize}
    \item \textbf{Integrating camera-control and semantic cues more cohesively.}
Our approach employs two separate modules to govern shot transitions: one conditioned on high-level semantic information and the other on parameter-level camera control signals. A compelling future direction is to investigate how to unify these two forms of conditioning more seamlessly, enabling a more coherent and expressive transition modeling process.
    \item  \textbf{Toward longer videos with more shot transitions.}
Extending our framework to generate longer videos that contain a richer set of shot transitions represents another valuable research direction. We believe that scaling to longer temporal horizons is feasible with additional data. Given that the effectiveness of our method has been verified on \datasetname, further fine-tuning on extended datasets may enable the model to generalize to videos with more shots and significantly longer durations.
\end{itemize}

\end{document}